\documentclass{article} % For LaTeX2e
\usepackage{iclr2026_conference,times}

\usepackage{graphicx}
\usepackage{booktabs}
\usepackage{multirow}
\usepackage{marvosym}
\usepackage{pifont}
\usepackage{xspace}

\usepackage{amsmath,amsfonts,bm}

\def\eqref#1{equation~\ref{#1}}
\def\1{\bm{1}}

\DeclareMathAlphabet{\mathsfit}{\encodingdefault}{\sfdefault}{m}{sl}
\SetMathAlphabet{\mathsfit}{bold}{\encodingdefault}{\sfdefault}{bx}{n}

\usepackage{hyperref}
\usepackage{url}

\newcommand{\ufo}{UFO$^{3}$\xspace}

\title{DevicesWorld: Benchmarking Cross-Device Agents in Heterogeneous Environments}

\iclrfinalcopy

\author{%
\begin{minipage}{\dimexpr\textwidth-2\tabcolsep\relax}
\vspace{2em}
\normalfont\raggedright
\textbf{Huatao Li}\textsuperscript{1}, \textbf{Xinwei Geng}\textsuperscript{2\,\Letter}, \textbf{Yuheng Wang}\textsuperscript{1}, \textbf{Yutong Li}\textsuperscript{1}, \textbf{Runde Yang}\textsuperscript{1}, \textbf{Hantao Chen}\textsuperscript{1},\\
\textbf{Shu Yao}\textsuperscript{1}, \textbf{Jingru Fan}\textsuperscript{1}, \textbf{Xuhui Ren}\textsuperscript{2}, \textbf{Yuanyuan Zhao}\textsuperscript{2}, \textbf{Fei Huang}\textsuperscript{2}, \textbf{Chen Qian}\textsuperscript{1\,\Letter}\\[0.4em]
\textsuperscript{1}School of Artificial Intelligence, Shanghai Jiao Tong University \qquad \textsuperscript{2}Honor Device Co., Ltd\\[0.4em]
\texttt{huataoli@sjtu.edu.cn}\qquad \texttt{gengxinwei@honor.com}\qquad \texttt{qianc@sjtu.edu.cn}\\
\centering
\vspace{0.6em}
\url{https://github.com/AgenticOrgLab/DevicesWorld}
\end{minipage}%
}

\begin{document}

\maketitle

{\renewcommand\thefootnote{}%
\footnotetext{\textsuperscript{\Letter}\,Corresponding authors.}}

% 修改页眉文字，并保留横线
\lhead{Preprint.}
\renewcommand{\headrulewidth}{0.4pt}

\begin{abstract}
LLM-based agents have rapidly improved at operating individual digital
environments such as mobile applications, desktop systems, and smart homes.
However, real-world user goals often span multiple devices: information may
come from a phone, be processed on a desktop, and the result may need to
appear on another device. Most existing benchmarks center on a single
dominant execution environment, making it difficult to evaluate whether
agents can acquire and integrate information across heterogeneous devices
and complete end-to-end tasks with cross-device dependencies. We introduce
\textsc{DevicesWorld}, a large-scale executable benchmark for cross-device
collaborative operation. \textsc{DevicesWorld} contains 6,140 tasks and
integrates three classes of device environments---mobile, desktop, and
IoT---into a unified cross-device interaction and evaluation framework.
Each task defines a natural-language user goal, participating devices and
initial states, executable actions, rule-based verifiers, and a cleanup
procedure. A multi-stage construction and quality-control pipeline keeps
tasks close to realistic user needs while allowing final outcomes to be
automatically verified from device states and generated files. We evaluate
five frontier LLM-agent systems on a fixed evaluation set. All methods
achieve low success rates, with the best reaching only 12.5\%. Among failed
runs, about 28.7\% satisfy at least one scoring condition yet still fail
the full task. Trajectories show that agents become stuck acquiring
information or manipulating interfaces, confuse source and output devices,
or terminate before all conditions are jointly satisfied. \textsc{DevicesWorld} turns cross-device collaborative operation into an
executable, reproducible, and diagnostically useful evaluation problem for
research on reliable cross-device agents.
\end{abstract}

\section{Introduction}
\begin{figure}[t]
\begin{center}
\includegraphics[width=\linewidth]{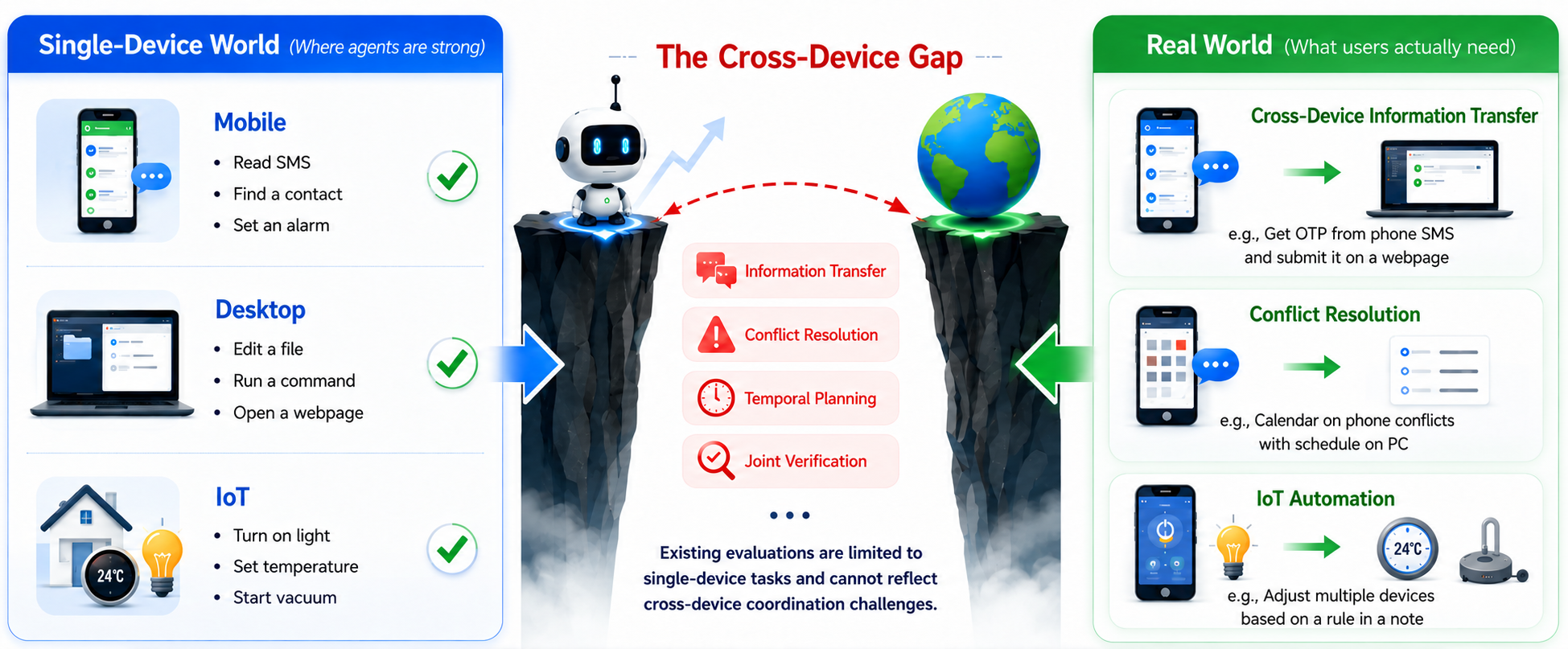}
\end{center}
\caption{Motivation and scope of \textsc{DevicesWorld}. Most existing
agent benchmarks center on single-device settings, whereas real-world
goals often require coordinated actions across multiple devices.
\textsc{DevicesWorld} bridges this gap with a large-scale executable
task suite and a unified cross-device interaction and evaluation
environment.}
\label{fig:teaser}
\end{figure}

Large language model–based agents are evolving from systems that generate responses into systems that directly operate digital environments. In recent years, LLM agents have become capable of performing relatively complex tasks in mobile applications, desktop systems, websites, and smart-home environments, demonstrating a degree of perception, planning, and execution capabilities ~\cite{rawles2025androidworld,xie2024osworld,
zhou2024webarenarealisticwebenvironment,koh2024visualwebarena,
li2025homebench,seo2025simuhome}. Real-world user tasks, however, are rarely confined to a single device. Phones store text messages, contacts, calendars, and location information; desktop systems contain documents, spreadsheets, websites, and code; and connected devices represented by smart-home systems provide environmental state, device-control, and automation capabilities. Completing a user goal often requires continuously acquiring information, processing data, and executing actions across these devices. Prior research on multi-device ecosystems has similarly highlighted the challenges introduced by heterogeneous device capabilities, changing device availability, and the distribution of information
and interface components across devices
~\cite{grubert2016multidevice,park2018adam}. Figure~\ref{fig:teaser} illustrates this gap.

This paper focuses on a class of interactive tasks that must be completed across multiple devices, which we refer to as cross-device collaborative operation. Given a natural-language user goal and a set of devices, an agent must acquire information from different devices over multiple rounds of interaction, determine which device should be used for the next action, and continue advancing the task until all required device states and outputs have been produced. For example, an agent may need to read a verification code from one phone, retrieve contact information from another phone, and then fill in and submit a form in a desktop browser. Alternatively, it may need to combine a phone calendar, a desktop spreadsheet, and the state of household IoT devices to create a scheduled task.

Such tasks are not equivalent to simply providing the model with more tools. Different devices expose incompatible observation formats, action interfaces, and state representations, and information obtained from one device often determines subsequent operations on another. At the same time, task outcomes may be distributed across several locations: a note may need to be written on a designated phone, a structured file may need to be saved on a particular desktop device, and a scheduled SmartHome operation may need to contain both the correct action and the correct time. Consequently, local progress on a single device does not directly imply that the overall task has been completed.

Most existing executable agent benchmarks still center on a single dominant execution environment. Mobile and GUI benchmarks mainly focus on operations within phones or graphical interfaces ~\cite{rawles2023androidwildlargescaledataset,lee2025benchmarkingmobiledevicecontrol,
rawles2025androidworld,lu2025guiodysseycomprehensivedatasetcrossapp}; desktop benchmarks primarily evaluate file, browser, command-line, and application interactions within a single computer environment ~\cite{xie2024osworld,bonatti2024windowsarena,
li2026windowsworld}; and smart-home benchmarks typically focus on state queries, device control, and automation within a single household environment ~\cite{li2025homebench,seo2025simuhome,li2026smhbench}. These benchmarks have substantially advanced research on agents in individual environments, but they generally cannot directly measure whether an agent can preserve information and device roles across multiple heterogeneous devices, organize a cross-device execution process, and ultimately satisfy completion conditions distributed across several devices.

One important reason for this evaluation gap is that constructing a cross-device benchmark is considerably more difficult than concatenating several single-device tasks. Each task must clearly specify the natural-language goal, participating devices, visible information sources, initial states, cross-device dependencies, target outputs, and verification conditions, while ensuring that these components remain mutually consistent. At the same time, the top-level evaluation environment must manage the initialization and state changes of multiple heterogeneous device environments, route the agent’s actions to the correct devices, and inspect multiple devices and outputs after execution. Tasks must also support repeatable initialization, automatic verification, and environment cleanup so that the benchmark can distinguish between an agent that merely produces a plausible plan and one that actually fulfills the user’s request. Comparable executable benchmarks bind task instructions to
deterministic initialization, state- or execution-based verification,
and reset or tear-down procedures
~\cite{rawles2025androidworld,xie2024osworld,
trivedi2024appworld}.

To address this gap, we introduce \textsc{DevicesWorld}, a large-scale executable benchmark for cross-device collaborative operation. \textsc{DevicesWorld} comprises a task suite of 6,140 tasks and a unified cross-device interaction and evaluation environment spanning three classes of device environments. Each task uses an executable specification to define the user goal, participating devices, initial states, and expected outcomes. The agent must perform actual operations in the environment, after which the system automatically determines whether the task has been completed based on the final device states and generated results. Tasks undergo multi-stage design, review, verification, and runtime testing to ensure executability and consistency between the stated requirements and the verification conditions.

We evaluate five representative LLM-agent baselines on \textsc{DevicesWorld}. All baselines achieve low task success rates, with the best result reaching only 12.5\%, indicating that current systems remain far from reliable cross-device execution. Trajectory analysis further shows that some failures occur after the agent has already made partial progress: the model may have read the required information, created a file, or completed part of a device-control process, yet still fail because other outputs are missing, device roles are confused, the final result is incorrect, or completion checks are insufficient. Approximately 28.7\% of failed runs satisfy at least one scoring condition without completing the full task. These results show that cross-device agents must not only perform local operations, but also continuously maintain cross-device dependencies, task progress, and the complete set of completion conditions.

The major contributions of this paper are summarized as follows:
\begin{itemize}
    \item \textbf{We construct a large-scale task set for cross-device collaborative operation.} The task set contains 6,140 unified executable task specifications spanning three classes of device environments, as well as diverse device combinations, application interfaces, file types, and task patterns. Tasks undergo multi-stage design, review, instantiation, verification, and runtime testing.
    \item \textbf{We develop a unified and executable cross-device interaction and evaluation environment.} The environment supports initialization, observation, target-device selection, action routing, state updates, result verification, environment cleanup, and trajectory recording across multiple heterogeneous devices, allowing agents to complete cross-device tasks within a single evaluation workflow. 
    \item \textbf{We systematically evaluate five representative LLM-agent baselines and conduct trajectory-level analysis.} The experiments reveal characteristic failures in required-information acquisition, target-side operation, error recovery, device-role maintenance, final-result generation, and task-completion checking. They also show that successful local operations do not naturally translate into complete cross-device task execution.
\end{itemize}

\section{Cross-Device Collaborative Operation}
\subsection{Problem Setting}
Following standard partially observable sequential-decision formulations and prior cross-environment agent modeling ~\cite{kaelbling1998pomdp,xu2024crab}, we formalize cross-device collaborative operation as an interactive execution task defined over an instance of a heterogeneous device environment. Let the set of device environments supported by
\textsc{DevicesWorld} be
\begin{equation}
    \mathcal{R} = \{\textit{Android},\ \textit{Linux},\ \textit{SmartHome}\}.
\end{equation}
Given a natural-language user goal $g$ and a set of devices
\begin{equation}
    \mathcal{D} = \{d_1, \ldots, d_n\}, \qquad \tau(d_i) \in \mathcal{R},
\end{equation}
where $\tau(d_i)$ denotes the environment type to which device $d_i$
belongs, each device $d_i$ has an internal state $s_i^{\,t}$ at step $t$.
The joint state of the cross-device environment is
\begin{equation}
    s^{t} = \bigl(s_1^{\,t}, \ldots, s_n^{\,t}\bigr).
\end{equation}
At the beginning of the task, the environment configures the device
states according to the task-specific initialization procedure:
\begin{equation}
    s^{0} = \mathrm{Init}_{\mathcal{T}}(\mathcal{D}).
\end{equation}
At step $t$, the agent receives the currently visible cross-device
observation
\begin{equation}
    O^{t} = \bigl\{\bigl(d_i,\ \tau(d_i),\
    \mathcal{O}_{\tau(d_i)}\bigl(s_i^{\,t}\bigr)\bigr)\bigr\}_{i=1}^{n},
\end{equation}
where $\mathcal{O}_{\tau(d_i)}$ is the observation function for the
corresponding device-environment type. Different device environments
may expose different observation formats, and an observation usually
reflects only the portion of the internal device state that is visible
to the agent.
 
Based on the user goal, current observations, and interaction history,
the agent selects a target device and an action:
\begin{equation}
    (d_t, a_t) \sim \pi\bigl(\cdot \mid g,\ O^{t},\ H^{t}\bigr),
    \qquad d_t \in \mathcal{D},\quad
    a_t \in \mathcal{A}_{\tau(d_t)}(d_t),
\end{equation}
where $\mathcal{A}_{\tau(d_t)}(d_t)$ denotes the action space supported
by the target device. After executing the action, the environment
updates the joint state:
\begin{equation}
    s^{t+1} = \mathcal{T}_{\tau(d_t)}\bigl(s^{t},\ d_t,\ a_t\bigr),
\end{equation}
and returns an action result or error signal $r_t$. The interaction
trajectory up to step $T$ is represented as the ordered sequence
\begin{equation}
    H^{T} = \bigl(\bigl(O^{t},\ d_t,\ a_t,\ r_t\bigr)\bigr)_{t=0}^{T-1}.
\end{equation}
After task execution ends, the task verifier checks the final device
states and generated outputs:
\begin{equation}
    (y, \rho) = \mathcal{V}_{\mathcal{T}}\bigl(s^{T},\ Z^{T}\bigr),
    \qquad y \in \{0, 1\},\quad \rho \in [0, 1],
\end{equation}
where $y$ indicates whether the task is fully successful, $\rho$ is a
partial score aggregated from the enabled scoring conditions, and
$Z^{T}$ represents task-relevant results generated or modified during
execution. A task is considered successful only when all required final
conditions are satisfied.
 
Cross-device collaborative operation therefore requires the agent to
decide both \emph{which device to operate} and \emph{which action to
execute} at every step, while ultimately ensuring that the states and
outputs of multiple devices jointly satisfy the user's goal.

\subsection{Why Cross-Device Operation is Challenging}
The difficulty of cross-device collaborative operation is not merely that it involves more steps. The agent must continuously maintain a unified understanding of the task objective, information sources, and completion state across multiple heterogeneous environments.

\paragraph{Observation formats and action interfaces are incompatible across devices.} Mobile, desktop, and IoT environments provide fundamentally different modes of interaction. Mobile tasks mainly involve applications, screenshots, and UI controls; desktop tasks may involve desktop GUIs, file systems, command-line interfaces, browsers, and document applications; and IoT environments operate through structured interfaces for querying devices, applying immediate controls, and creating scheduled operations. An agent therefore cannot use a single uniform interaction pattern for every device. It must interpret different forms of observation according to the current objective and map the high-level user request into executable actions in the appropriate device environment. Prior work on multi-device interfaces and multi-platform GUI agents similarly identifies heterogeneous device capabilities, interface divergence, grounding, and cross-platform planning as central challenges
~\cite{grubert2016multidevice,park2018adam,
wang2025mmbenchgui}.

\paragraph{Task-critical information is distributed across devices.} The information required to complete a task is usually not presented all at once in the current observation. One device may provide a verification code, another may store a contact, and a third may be responsible for submitting a form. Alternatively, a phone may provide calendar information, a Linux file may define a policy, and the SmartHome state may determine whether an automation can be executed. The agent must remember which device each piece of information came from, how that information will be used in later steps, and which target device should receive the result. The challenge therefore involves not only reading information, but also preserving, combining, and using information across devices. CRAB models cross-environment information transfer through composed
subtasks, while \ufo represents inter-device data and control dependencies explicitly in a distributed task graph ~\cite{xu2024crab,zhang2025ufo3}.

\paragraph{The final objective is jointly determined by distributed outcomes.} Cross-device tasks often contain several results that must all be satisfied simultaneously. For example, the agent may need to save a note on a designated phone, generate a structured file on a Linux device, and create a complete scheduled task in SmartHome. A missing output, an output written to the wrong location, or an inconsistency between results on different devices causes the entire task to fail. The agent must therefore maintain global task progress rather than deciding that the task is complete solely because the most recent local operation succeeded. Related executable benchmarks similarly evaluate task completion from multiple application states or jointly satisfied postconditions,
rather than from local action success alone
~\cite{trivedi2024appworld,xu2024crab,wei2026opencomputer}.

\section{DevicesWorld}

\subsection{Benchmark Overview}

\textsc{DevicesWorld} is a large-scale executable benchmark for coordinated operation across multiple devices. It is designed to evaluate whether LLM agents can complete end-to-end tasks in heterogeneous cross-device environments comprising mobile devices, desktop systems, and IoT devices. \textsc{DevicesWorld} contains 6,140 tasks, covering cross-device tasks within a single class of device environment, cross-device workflows spanning two classes of device environments, and complex tasks jointly involving Android, Linux, and SmartHome. A single task may involve up to four concrete device instances.

\textsc{DevicesWorld} covers a broad range of interaction surfaces. On Android, it includes common applications for text messaging, contacts, calendars, files, notes, music, maps, cameras, and audio recording; on Linux, it includes text and structured files, spreadsheets and documents, browsers, command-line tools, code projects, email drafts, and multimedia processing. SmartHome is a home-IoT simulation environment developed for \textsc{DevicesWorld}. It supports device-state queries, immediate control, and scheduled tasks, and includes 11 categories of devices: lights, dimmable lights, curtains, air conditioners, heaters, air purifiers, humidifiers, dehumidifiers, washing machines, dryers, and robotic vacuum cleaners.

The tasks cover a variety of cross-device workflow patterns, including direct information transfer, information extraction and transformation, multi-source integration, cross-device state synchronization, conflict resolution, conditional decision-making, infeasible-request reporting, output consistency across multiple endpoints, and device-state control. For example, an agent may need to obtain a verification code and contact information from two different phones and then submit a form in a Linux browser; it may need to determine whether to execute a device-control operation based on a Linux policy file and the current SmartHome state; and more complex tasks may require the simultaneous creation of an Android note, a Linux JSON file, and a SmartHome scheduled task.

The defining property of \textsc{DevicesWorld} is its executability. Each task corresponds to an interactive episode that can be initialized and run, rather than a static question. This design follows the executable-evaluation paradigm established by
recent web, mobile, desktop, and application benchmarks, in which task completion is determined from environment state rather than from a text-only answer
~\cite{zhou2024webarenarealisticwebenvironment,rawles2025androidworld,
xie2024osworld,trivedi2024appworld}. The agent must modify device states through actual actions, and the task outcome is checked by rule-based verifiers. The task suite spans multiple difficulty levels, ranging from relatively direct information handoffs between two devices to long-horizon tasks requiring multi-source integration, conditional decision-making, and consistency across outputs on multiple endpoints.

\subsection{Cross-Device Environment}

\textsc{DevicesWorld} uses a unified cross-device environment controller to manage Android, Linux, and SmartHome environments. For each task, the controller creates the required device instances according to the device configuration and handles initialization, action dispatch, state updates, evaluation, and cleanup. Every action produced by the agent specifies a target device. The environment checks whether the action belongs to the action space supported by the target device’s environment and routes it to that device for execution.

The Android environment is implemented on top of AndroidWorld ~\cite{rawles2025androidworld} and provides screenshots, UI elements, and GUI operations for mobile applications. The Linux environment is implemented on top of OSWorld ~\cite{xie2024osworld} and supports desktop GUIs, command-line operations, files, browsers, and common desktop applications. The SmartHome environment supports device queries and control, time advancement, scheduled operations, and result reporting. Unlike the Android and Linux environments, the complete SmartHome state is not automatically included in every observation; the agent must explicitly query the relevant device states and capabilities. The use of explicit state queries, time advancement, and scheduling
is related to recent executable smart-home environments
~\cite{seo2025simuhome,li2026smhbench}.

\subsection{Task Construction}

Constructing a cross-device task involves more than generating a natural-language instruction. A valid task must jointly specify the participating devices, visible information sources, initial environment states, executable operations, target outcomes, verification conditions, and cleanup procedure, while maintaining consistency across these components. A task that appears semantically plausible is not necessarily a valid executable benchmark instance: required information may not have been placed in the environment; the task may rely on operations unsupported by the corresponding device runtime; a resource file may have the correct extension yet remain unreadable by the intended application; or a verifier may enforce hidden requirements that are absent from both the user instruction and the observable environment. In addition, setup must establish only the initial task state rather than completing the target outcome in advance, while cleanup must remove task-induced states to avoid affecting subsequent episodes. 

\begin{figure}[t]
\begin{center}
\includegraphics[width=\linewidth]{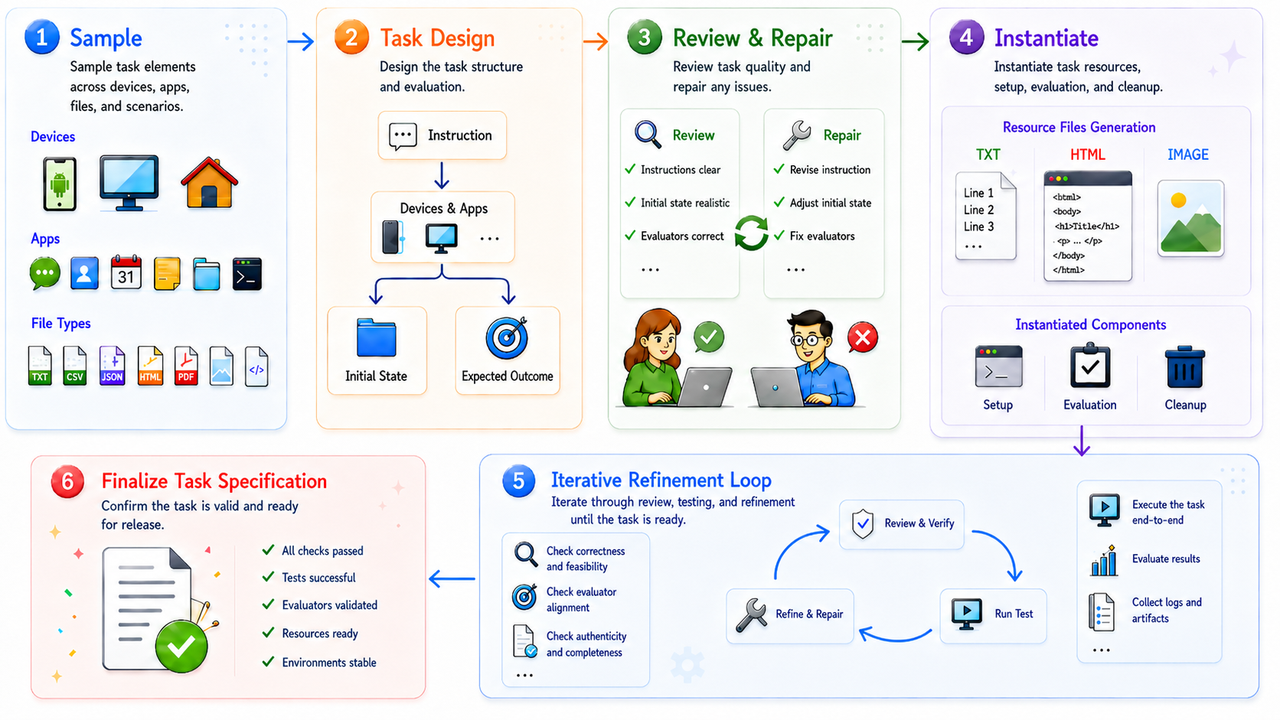}
\end{center}
\caption{The multi-stage task-construction and quality-control
pipeline of \textsc{DevicesWorld}. The pipeline samples a task
configuration, converts it into a concrete task design, reviews and
repairs the design, instantiates task resources and executable
components, and iteratively validates the candidate through verifier
checks and runtime testing before finalizing it as an executable task
specification.}
\label{fig:pipeline}
\end{figure}

Accordingly, cross-device task construction must satisfy three requirements. First, all information required to complete a task must be obtainable from either the instruction or an environment accessible to the agent. Task difficulty should arise from cross-device search, information transformation, conditional reasoning, and execution, rather than from missing prerequisite information. Second, every required operation and state transition must be supported by the actual interface of the corresponding device environment. Third, the task requirements, expected outcomes, and verifiers must remain aligned, such that a valid successful execution passes verification while missing or incorrect results do not. We organize these interdependent requirements into a task contract that is complete, executable, and verifiable.

To satisfy these requirements, we adopt the multi-stage task-construction and quality-control pipeline illustrated in Figure~\ref{fig:pipeline}. The pipeline begins by sampling the task space over the number and combination of devices, the applications or interaction surfaces involved, the input and output file types, the task scenario, and the feasibility conditions. For example, a sample may contain two Android devices and one Linux device, with text messages, contacts, and a browser form specified as the relevant interaction surfaces. This stage controls the coverage and diversity of the task suite but does not directly instantiate a concrete user instruction or expected outcome.
    
During task design, an LLM constructs a concrete task from the sampled configuration and the capabilities of the corresponding device environments. The design specifies the natural-language user goal, participating devices, visible information sources, required transformations or decisions, target outputs, and expected outcomes. The central information dependency can be expressed as

\[
\text{visible sources} \rightarrow \text{transformation or decision} \rightarrow \text{required outputs} \rightarrow \text{expected outcomes}.
\]

Every verified outcome must be derivable from information visible to the agent. The initial design then undergoes an independent review-and-repair stage, which examines whether the user request is natural, whether the specified devices genuinely participate in the task, whether the information sources are sufficient, whether the required operations are supported by the device environments, and whether the instruction is consistent with the evaluation criteria. Designs with identifiable defects are revised accordingly, whereas candidates that cannot be made internally consistent are discarded before instantiation.

After semantic review, the task design is instantiated as a complete executable task specification. For required resources—including text, CSV, JSON, webpages, Office documents, PDFs, images, audio, video, archives, and code projects—the LLM specifies the semantic content and structure, while deterministic programs generate format-valid files that can be opened by the intended applications. The system then instantiates setup, evaluation, and cleanup. Setup deploys resource files and source states to the designated devices without generating the target outcome in advance; evaluation constructs verifiers from the expected outcomes defined during task design; and cleanup removes files, application data, and SmartHome states created or modified by the task. The instruction, devices, setup, evaluation, cleanup, and execution limits are then assembled into a unified task specification.

Automatically assembled tasks remain candidate instances and enter an iterative refinement loop consisting of deterministic validation, runtime testing, and targeted repair. Deterministic validation checks the task schema, compatibility between devices and interfaces, file paths, resource-file formats, verifier coverage, and whether setup inadvertently satisfies the task objective. For interactions that are sensitive to runtime conditions, we additionally test whether initialization executes correctly, whether the intended outcome is recognized by the verifier, whether missing or incorrect outcomes fail verification, and whether cleanup removes the principal task states. When validation or testing identifies a problem, the relevant instruction, resource files, setup, evaluation, or cleanup component is revised, and the corresponding checks are rerun. Candidates that repeatedly fail these checks are excluded from the benchmark. Verifier validation is particularly important because rule-based
evaluators may fail to recognize some successful trajectories ~\cite{lu2025agentrewardbench}. Recent verifier-grounded frameworks therefore use execution feedback and application-state inspection to improve evaluation reliability and auditability
~\cite{wei2026opencomputer}. This procedure corresponds to the Review \& Verify, Run Test, and Refine \& Repair loop in Figure~\ref{fig:pipeline}.

LLMs, deterministic programs, and release review serve distinct roles throughout the pipeline. LLMs are primarily responsible for designing user scenarios, task semantics, and cross-device information dependencies. Deterministic programs generate resource files, assemble task specifications, and check schemas, interfaces, paths, file formats, and verifier coverage. Release review focuses on the naturalness of the instruction, the alignment between task requirements and verification conditions, and the executability of the task in the actual device environments. Only candidates that pass the required review, validation, testing, and retesting procedures are finalized as official task specifications.

\textsc{DevicesWorld} combines this unified multi-stage construction pipeline with task-specific builders and manually designed tasks. Regardless of their origin, all tasks are converted into the same task-specification format and undergo the review, validation, and runtime testing appropriate to their task type and interaction surface.

\subsection{Task Execution and Evaluation}

Task execution in \textsc{DevicesWorld} is managed by a unified cross-device
environment, as illustrated in Figure~\ref{fig:execution}. Each task
consists of five stages:

\begin{figure}[t]
\begin{center}
\includegraphics[width=\linewidth]{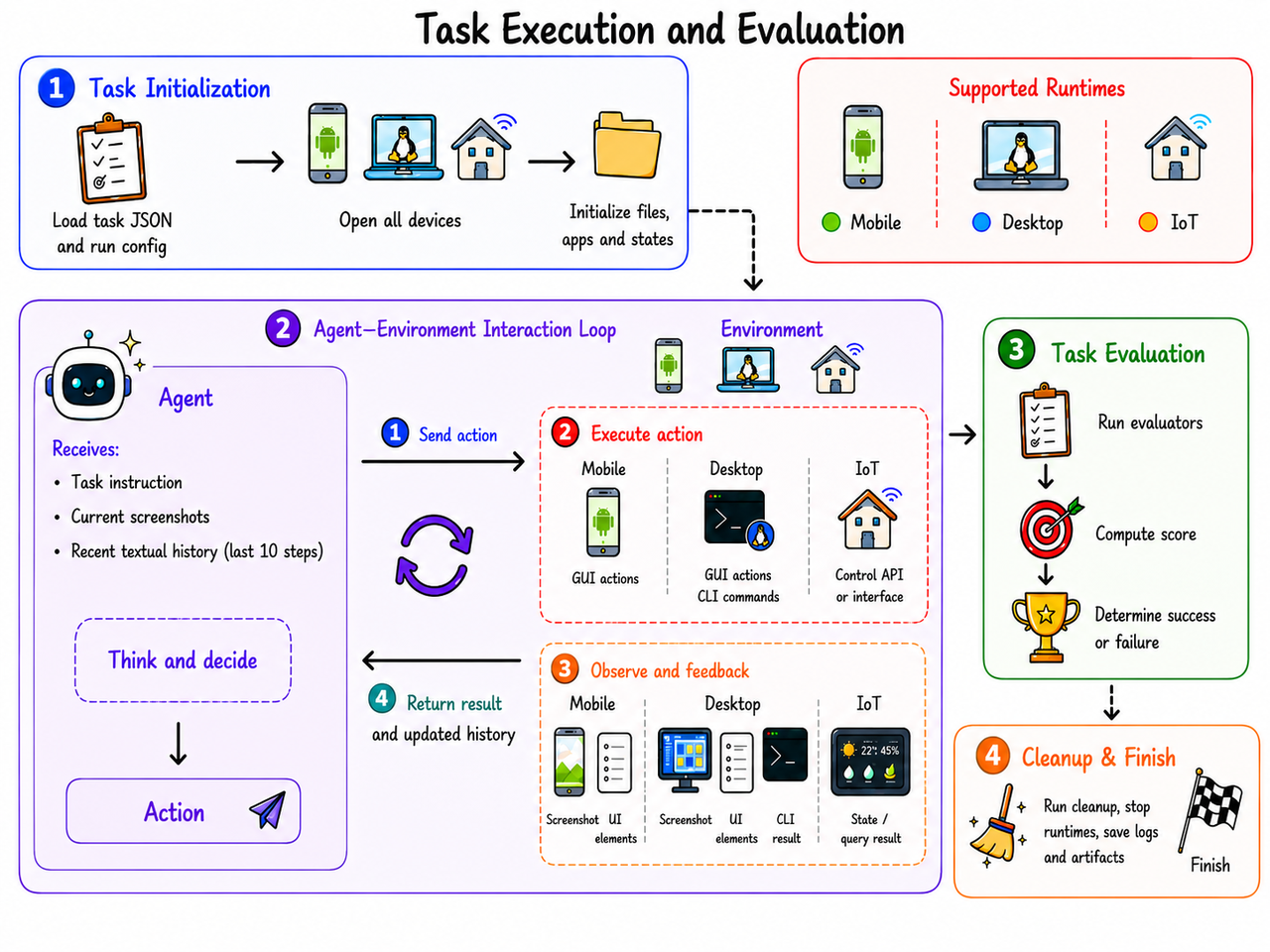}
\end{center}
\caption{Task execution and evaluation workflow in
\textsc{DevicesWorld}. The environment first initializes all
participating mobile, desktop, and IoT devices. During interaction,
the agent repeatedly selects a target device and issues an
environment-specific action based on the current observations and
execution feedback. After termination, task-specific evaluators
compute the partial score and strict success outcome, followed by
cleanup of task-induced states and artifacts.}
\label{fig:execution}
\end{figure}

\begin{itemize}
    \item \textbf{Initialization} The environment configures the relevant device environments according to the task setup, for example by inserting Android application data, preparing Linux files and webpages, or setting initial SmartHome device and automation states. 
    \item \textbf{Observation} The agent receives the currently visible cross-device observations. Android and Linux may expose screenshots, UI elements, files, or terminal-related information, whereas SmartHome state must be actively obtained through query interfaces. 
    \item \textbf{Action execution} The agent selects a target device and outputs an action supported by that device environment. The environment executes the action, updates the internal state, and returns a new observation and feedback. 
    \item \textbf{Evaluation} After the task ends, the system runs task-specific evaluators to inspect final device states, application states, browser states, file contents, and other relevant outputs. 
    \item \textbf{Cleanup} After evaluation, the environment removes states and outputs introduced by the task, reducing residual state across different task episodes. 
\end{itemize}

This process requires the agent to complete the task through actual actions rather than merely describing a plan. \textsc{DevicesWorld} also records step-by-step trajectories and evaluator traces. It can therefore report final success while supporting analysis of where the model became stuck, selected the wrong device, or terminated before the final conditions were satisfied.

\textsc{DevicesWorld} primarily uses rule-, state-, and file-based checks, in line with the programmatic end-state evaluation paradigm adopted by executable agent benchmarks ~\cite{rawles2025androidworld,trivedi2024appworld}. On Linux, verifiers can inspect files, structured data, and browser states; on Android, they can inspect text messages, contacts, calendars, notes, and application data; and in SmartHome, they can inspect device states, existing scheduled-task information, and feasibility reports. A task succeeds only when all required final conditions are jointly satisfied, whereas the mean score reflects the proportion of enabled scoring conditions that have been satisfied.

\section{Experiments}
\subsection{Experimental Setup}

We select a fixed evaluation set from the \textsc{DevicesWorld} task suite using a coverage-driven stratified selection strategy. The selection aims to cover major device combinations, representative applications and interaction surfaces, task types, and difficulty levels. We also prioritize tasks with natural user instructions, clearly defined information sources, repeatable initialization, and automatically verifiable outcomes. All baselines are evaluated on the same task set, runtime environments, and verification conditions.

We evaluate five baselines: GPT-5.5, Qwen3.7-Plus, Gemini-3.1-Pro-Preview, Claude Opus 4.8, and \ufo. The first four baselines use the same direct-interaction protocol and context configuration, differing only in the underlying model. \ufo is a hierarchical cross-device agent framework ~\cite{zhang2025ufo3}. In our setup, \ufo uses GPT-5.5 as its underlying model, decomposes each user task into device-level subtasks, and uses an adapter to convert its outputs into the target-device and action format required by \textsc{DevicesWorld}.

Each task permits at most 50 interaction steps. At every step, the four direct LLM baselines receive the task instruction, the currently visible cross-device observations and screenshots, the previous action and its execution feedback, and the textual interaction history from the most recent ten steps. The model then outputs an executable action on a target device. The decoding temperature is set to 0. The complete internal state of SmartHome is not automatically provided to the agent; the model must use the corresponding query actions to obtain the required device states, capabilities, and information about existing scheduled tasks.

We report the following metrics:

\begin{itemize}
    \item \textbf{Task Success Rate} The proportion of tasks for which all required verification items pass. This is a strict task-level metric: if any required condition is not satisfied, the task is considered unsuccessful.
    \item \textbf{Mean Score} Each task contains one or more scored evaluation points. For a task with $K_i$ scored evaluation points, each point contributes $1/K_i$ to the task score. The task score is the sum of the contributions from all satisfied points, and Mean Score is the average task score across all tasks.
    \item \textbf{Average Steps} The average number of interaction steps executed per task.
    \item \textbf{Average Duration} The average end-to-end runtime per task. 
    \item \textbf{Budget Exhaustion Rate} The proportion of tasks that reach the interaction limit without being successfully completed. 
    \item \textbf{Average Tokens per Task} The average number of input and output tokens used per task under a unified counting protocol. 
\end{itemize}

% =========================================================
% Sections 4.2 (Results) and 4.3 (Failure Analysis)
% ICLR style. Requires in preamble: \usepackage{graphicx}
% Figures: ./Figures/4.png (termination modes), ./Figures/5.png (failure heatmap)
% =========================================================

\subsection{Results}
\label{subsec:results}

Table~\ref{tab:main_results} summarizes the overall performance and
execution statistics of the five baselines on \textsc{DevicesWorld}.
All five baselines achieve low task success rates, ranging from 9.5\%
to 12.5\%. \ufo with GPT-5.5 achieves the highest success rate at
12.5\%, while GPT-5.5 and Gemini-3.1-Pro-Preview both reach 12.0\%. The
differences in success rate among the baselines are relatively small.
The central conclusion is therefore not that one system significantly
outperforms the others, but that current agents generally struggle to
complete cross-device tasks reliably.

\begin{table}[t]
\caption{Overall task-completion and execution statistics on the
fixed \textsc{DevicesWorld} evaluation set. Success and Mean Score measure strict and partial task completion, respectively, while the remaining columns summarize execution behavior. Budget Exh. is computed over all evaluated tasks. Bold indicates the best value in each comparable column.}
\label{tab:main_results}
\begin{center}
\resizebox{\textwidth}{!}{%
\begin{tabular}{lcccccc}
\toprule
\multirow{2}{*}{\bf Method}
& \multicolumn{2}{c}{\bf Task Completion}
& \multicolumn{4}{c}{\bf Execution Statistics} \\
\cmidrule(lr){2-3}\cmidrule(lr){4-7}
& {\bf Success (\%)}$\,\uparrow$ & {\bf Mean Score}$\,\uparrow$
& {\bf Steps} & {\bf Dur.\ (min)}
& {\bf Budget Exh.\ (\%)}$\,\downarrow$ & {\bf Tok./Task}$\,\downarrow$ \\
\midrule
GPT-5.5                & 12.0       & {\bf 0.262} & {\bf 22.1} & 7.9  & {\bf 22.0} & 413k \\
Qwen3.7-Plus           & 9.5        & 0.201       & 33.5 & {\bf 4.5}  & 56.0       & 606k \\
Gemini-3.1-Pro-Preview & 12.0       & 0.252       & 26.0 & 6.6  & 33.5       & 406k \\
Claude Opus 4.8        & 10.5       & 0.213       & 26.0 & 12.0 & 35.0       & 561k \\
\midrule
\ufo                   & {\bf 12.5} & 0.212       & 25.4 & 11.5 & 38.5       & {\bf 278k} \\
\bottomrule
\end{tabular}%
}
\end{center}
\end{table}

Success rate and mean score provide complementary perspectives. Success
rate requires all necessary verification conditions defined by a task
to be satisfied, whereas mean score reflects the average satisfaction
of score-enabled conditions. Across the five baselines, approximately
28.7\% of failed runs satisfy at least one scoring condition. This
indicates that a substantial fraction of failures occur after the agent
has already made partial progress rather than producing no useful
result at all. An agent may have read information from a device,
generated a Linux file, or completed a SmartHome control operation, yet
still fail because another output is missing, the result is written to
the wrong device, its content or structure is incorrect, or the final
state remains incomplete.

For tasks that jointly involve Android, Linux, and SmartHome, none of
the five baselines successfully completes a task. When information
sources, execution actions, and final outcomes are distributed across
all three classes of device environments, current agents struggle to
maintain a stable end-to-end execution process. This result, however,
should not be interpreted solely as a coordination gap, because some
failures also involve mobile-GUI localization, target-side operations,
structured-output generation, and result persistence on a single
device.

In addition to task-completion metrics,
Table~\ref{tab:main_results} reports execution length, runtime, budget
exhaustion, and token usage. Budget Exhaustion Rate is the proportion
of all evaluation tasks for a baseline whose execution trajectory ends
because the interaction-step or time budget is reached without
successful completion. Average Tokens per Task aggregates the average
token usage during task execution under a unified counting protocol
and describes the context burden introduced by long-horizon
cross-device interaction. \ufo's Average Steps value records actual
environment actions, while its execution process also includes planner
and device-agent decision rounds. This metric is therefore primarily
descriptive and should not be used for a strict single-variable
comparison of architectures. Average Duration is similarly affected by
model-service latency and device-response speed.

The execution statistics show that more interactions or greater token
usage do not naturally translate into better task performance.
Qwen3.7-Plus uses the largest number of average steps and tokens and
has the highest budget-exhaustion rate, yet its success rate and mean
score are both the lowest among the five baselines. Longer trajectories
do not necessarily represent sustained effective progress; they may
contain repeated searches, stalled interface operations, or execution
strategies that are not adjusted in time. \ufo uses the fewest average
tokens and achieves a slightly higher success rate than the other
baselines, but its mean score does not improve correspondingly, and
budget exhaustion remains substantial.

Overall, the limitations of current systems are not simply a lack of
available steps or context. The more important issue is whether agents
can use existing observations effectively, revise their plans after
becoming blocked, and continuously maintain the conditions that remain
incomplete. The specific termination modes and execution problems are
examined further in Section~\ref{subsec:failure_analysis}.

\subsection{Failure Analysis}
\label{subsec:failure_analysis}

Task-specific verifiers can determine whether the final device states
and generated results satisfy the user goal, but they cannot directly
explain \emph{how} an agent failed. We therefore analyze failures at
three levels using execution trajectories, action feedback, and
evaluator traces. We first identify how failed trajectories terminate.
We then assign each trajectory to one primary failure type according to
the earliest major deviation that dominates the subsequent failure.
Finally, we examine representative trajectories to investigate the
likely execution mechanisms underlying these failure types.

\subsubsection{Two Termination Modes for Failed Runs}
\label{subsubsec:termination_modes}

\begin{figure}[t]
\begin{center}
\includegraphics[width=0.85\linewidth]{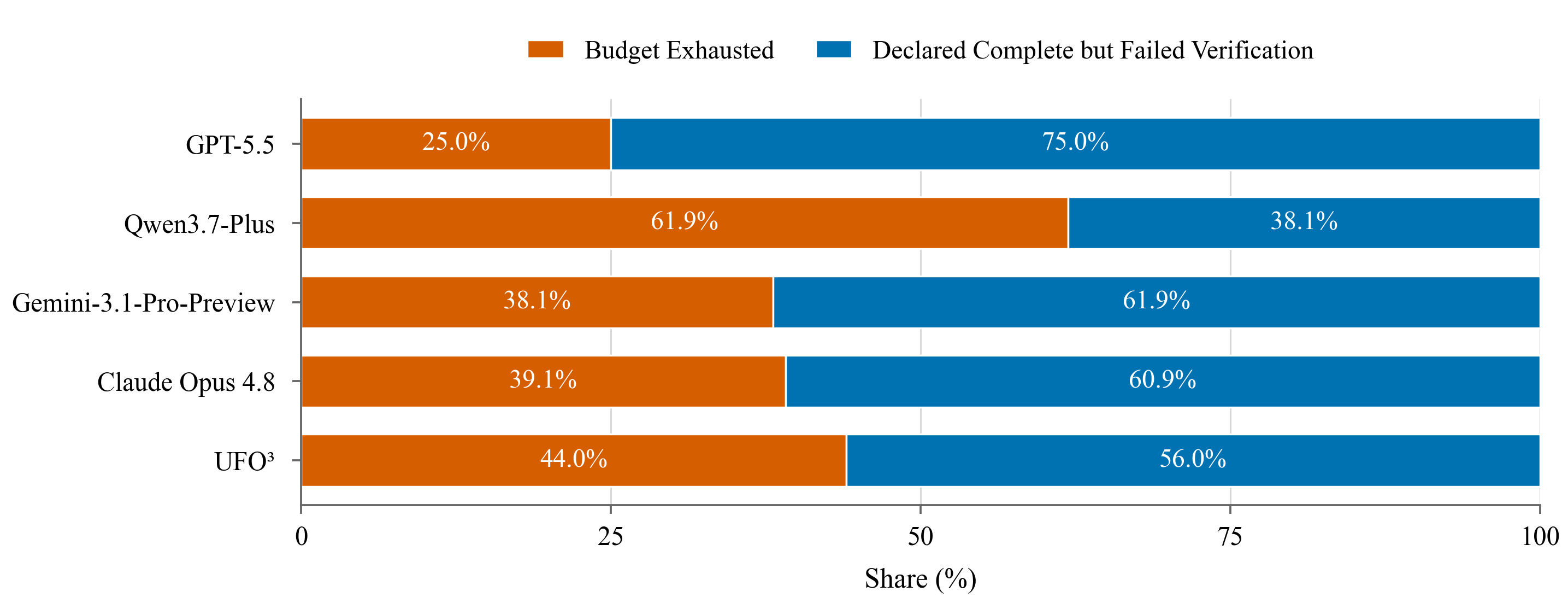}
\end{center}
\caption{Termination modes among failed runs. For each baseline,
failed trajectories are partitioned into those that exhaust the
interaction budget and those in which the agent declares completion
but fails final verification. Percentages are normalized over the
failed runs of each baseline and sum to 100\%.}
\label{fig:termination_modes}
\end{figure}

Figure~\ref{fig:termination_modes} shows how failed runs terminate for
each baseline. Overall, 41.7\% of failed runs exhaust the interaction
budget before completing the task, while the remaining 58.3\% are
terminated by the agent declaring the task complete even though the
final state fails verification. The percentages are calculated
separately over each baseline's failed runs. The two categories are
mutually exclusive and jointly cover all failed runs.

The baselines exhibit clearly different failure-termination
distributions. Qwen3.7-Plus is more likely to continue operating
without converging, with budget exhaustion accounting for 61.9\% of its
failed runs. GPT-5.5 more frequently terminates before all final
conditions are satisfied, with 75.0\% of its failed runs ending in a
completion declaration that does not pass verification.
Gemini-3.1-Pro-Preview, Claude Opus 4.8, and \ufo also more often
declare completion and subsequently fail verification, although budget
exhaustion still accounts for 38.1\%, 39.1\%, and 44.0\% of their
respective failed runs.

\begin{figure}[t]
\begin{center}
\includegraphics[width=\linewidth]{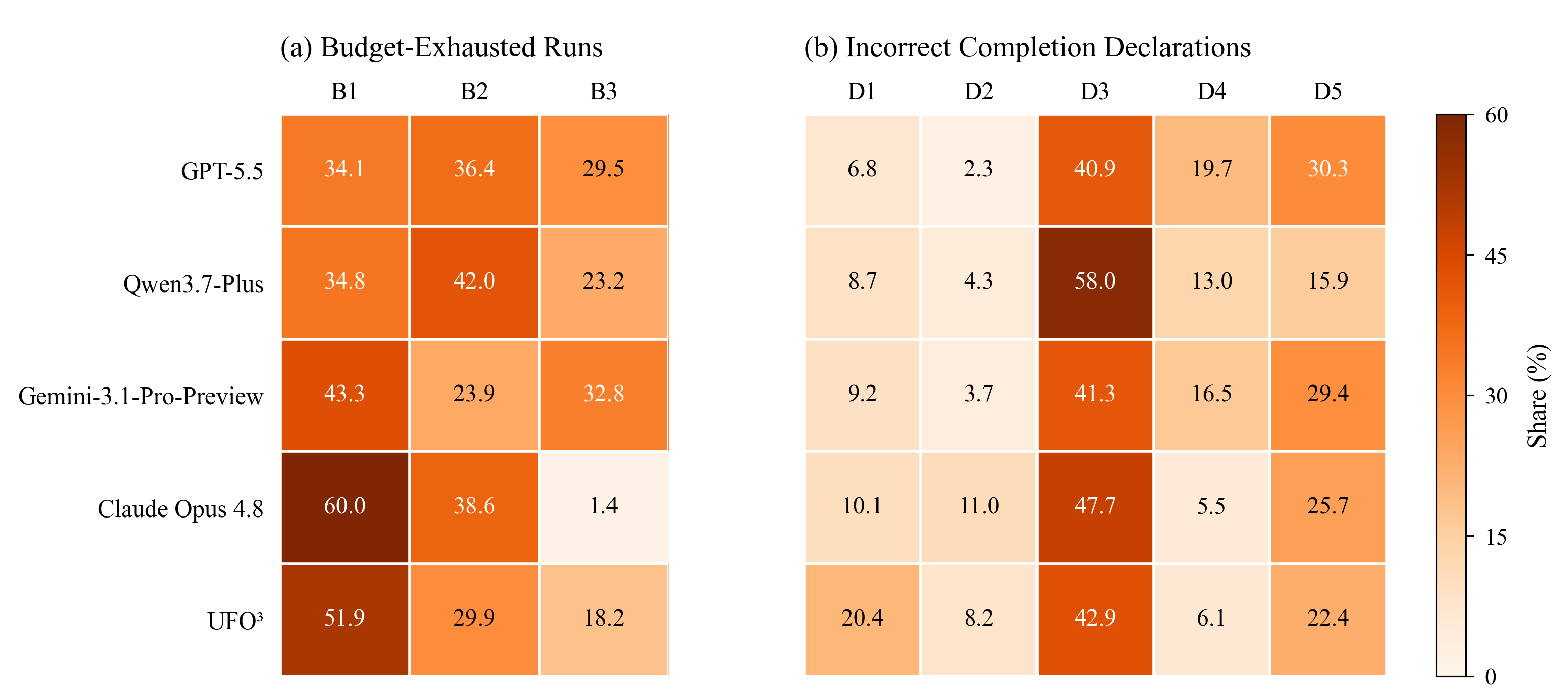}
\end{center}
\caption{Primary failure-type distributions within the two
termination modes. Panel (a) categorizes budget-exhausted runs into
failure to acquire required source information (B1), failure to
complete the target-side operation (B2), and failure to recover after
an explicit execution error (B3). Panel (b) categorizes incorrect
completion declarations into omission of a required subgoal (D1),
confusion about device roles or target location (D2), incorrect final
output or device state (D3), completion without confirming result
persistence (D4), and incomplete satisfaction of final conditions
across multiple devices (D5). Each row is normalized to 100\% within its corresponding panel.}
\label{fig:failure_types}
\end{figure}

These results show that failures do not concentrate in a single
termination mode. One class of failures consists of continued
interaction without effective convergence, while the other consists of
an agent ending execution even though the final state does not satisfy
the task conditions. Termination mode alone, however, explains only how
a task ends, not where the execution chain first deviates.
Figure~\ref{fig:failure_types} therefore breaks down the main failure
types within the two termination modes.
Figure~\ref{fig:failure_types}(a) corresponds to budget-exhausted
trajectories, and Figure~\ref{fig:failure_types}(b) corresponds to
trajectories in which the agent declares completion but fails
verification. Each trajectory is assigned to exactly one primary
failure category within its termination mode, so each row in the
corresponding heatmap sums to 100\%.

\subsubsection{Major Failure Types in Budget-Exhausted Runs}
\label{subsubsec:budget_failures}

Figure~\ref{fig:failure_types}(a) divides budget-exhausted trajectories
into three categories: failure to acquire required source information,
failure to complete the target-side operation, and failure to recover
after an explicit execution error. Across all five baselines, the three
categories account for 44.6\%, 34.9\%, and 20.5\% of budget-exhausted
runs, respectively. Claude Opus 4.8 and \ufo have budget-exhausted
failures concentrated more heavily in required-source-information
acquisition; Qwen3.7-Plus has the highest proportion of target-side
operation failures; and Gemini-3.1-Pro-Preview has a comparatively high
proportion of explicit-error recovery failures.

\paragraph{Failure to acquire required source information (B1).}
These tasks require the agent to first obtain information from a source
device, application, or file before a later operation can be performed.
Examples include a calendar time, a contact, a map location, a
playlist, or policy content. The model often selects the correct
information source and may even enter the correct application, but
repeatedly fails to open the page containing the target value or to
read the complete information from the current interface. Typical
trajectories involve repeatedly switching among different Calendar
views, repeatedly searching in Files, OsmAnd, or Markor, or remaining
on a page that does not contain the target information.

These failures are often more than a single incorrect click or search.
Empty search results, unchanged interfaces, or persistent dialog boxes
may already be visible in the observations, but the agent fails to
derive a new search strategy from that feedback. It also fails to
explicitly track which information has been confirmed and which remains
missing. When the missing information is a prerequisite for an
operation on Linux, SmartHome, or another phone, a local
state-acquisition problem blocks the entire cross-device execution
chain. For example, in a calendar-conflict task, an agent may have
already read the authoritative policy and original spreadsheet on Linux
but remain unable to obtain the correct time from Android Calendar. As
a result, it never begins the subsequent Linux file update.

B1 therefore exposes not only GUI-localization difficulty, but also
weak recognition of execution stagnation. The agent cannot determine
that the current search path has failed to produce new information for
an extended period, nor can it switch to an alternative interaction
strategy in time. B1 accounts for 60.0\% of Claude Opus 4.8's
budget-exhausted trajectories and 51.9\% of \ufo's, indicating that a
large fraction of their budget failures occur at the source-information
stage of a cross-device dependency chain.

\paragraph{Failure to complete the target-side operation (B2).}
In these trajectories, the agent has already acquired the required
information or has clearly entered the output stage, but cannot
complete creation, editing, saving, sending, submission, scheduling, or
state modification on the target device. For example, the model may
know what a note should contain but remain stuck in a file-creation
dialog, or it may have filled in a web form but be unable to confirm
submission.

This category exposes the gap between \emph{knowing what should be
done} and \emph{actually changing the environment state}. Agents often
cannot reliably handle input focus, time pickers, save buttons, and
multi-stage dialogs. They also lack explicit post-operation checks.
After clicking Save or Send, an agent may not immediately verify that
the result actually exists. When the interface remains unchanged for a
long time, it may fail to switch input methods, reopen the result, or
use an alternative interaction path.

In hierarchical agents, the same problem can arise between the planner
and the device agent. The planner may have completed the upstream
reasoning but transmit incomplete concrete values or output
requirements to the target-device agent. The device agent then knows
the general type of operation it should perform but cannot produce the
exact required result. Qwen3.7-Plus has the highest B2 proportion among
its budget-exhausted trajectories, at 42.0\%, indicating that many of
its long trajectories reach the execution stage but still fail to
produce the required target-side output.

\paragraph{Failure to recover after an explicit execution error (B3).}
This category requires that the trajectory contain an explicit action,
protocol, command, or runtime error and that the agent fail to develop
an effective alternative after receiving the error feedback. Typical
cases include invalid UI indices, incorrect target devices,
command-syntax or encoding errors, missing valid actions, and repeated
action-execution timeouts.

An explicit error does not necessarily have to cause task failure. The
key question is whether the agent can use the feedback to change its
strategy. In current trajectories, models frequently make only a minor
modification to the previous action and then trigger the same or a
similar error again, rather than switching interfaces, operation paths,
or target devices. This suggests that although error information
appears in the subsequent context, it is not consistently converted
into replanning and recovery behavior.

B3 includes both model-side errors and runtime or tool instability.
Invalid actions, incorrect devices, and stale UI indices more directly
reflect problems in action generation or grounding, whereas repeated
Android action timeouts may also involve adapters or the runtime
environment. B3 should therefore be interpreted as a failure to
establish effective recovery after an explicit error, rather than being
attributed entirely to model reasoning.

Overall, budget exhaustion usually develops along a continuous failure
chain. Acquisition of required source information or a target-side
operation is first blocked. The agent then repeatedly makes local
attempts without recognizing execution stagnation or recovering
effectively from feedback. The interaction budget becomes concentrated
on one device or subgoal, preventing later device operations from
beginning, until the step or time limit is reached. Budget exhaustion
therefore does not imply that the task inherently requires more steps.
Instead, it reflects deficiencies in progress monitoring, strategy
switching, error recovery, and the organization of cross-device
execution.

\subsubsection{Major Failure Types in Incorrect Completion Declarations}
\label{subsubsec:completion_failures}

Figure~\ref{fig:failure_types}(b) divides trajectories in which the
agent declares completion but fails final verification into five
categories: omission of a required subgoal, confusion about device
roles or target location, an incorrect final output or device state,
declaring completion without confirming result persistence, and
incomplete satisfaction of final conditions across multiple devices.
Incorrect final output or device state is the most common category
across all five baselines, accounting for 45.1\% overall and
representing the largest category for every baseline. Incomplete
satisfaction of cross-device final conditions is the second most
common, accounting for 25.7\%.

\paragraph{Omission of a required subgoal (D1).}
Before invoking the completion action, the agent never actually
executes a required device operation, generates a required file, sends
a required reply, modifies a required state, or produces another
required deliverable. For example, the model may complete the main JSON
file or spreadsheet but never open the required webpage, send the
required text message, write the required note, or perform the required
SmartHome operation.

This failure indicates that the corresponding subgoal was never stably
incorporated into the task plan or disappeared from the set of
remaining objectives during a long execution. For direct LLM baselines,
this usually appears as overemphasis on the most salient primary output
in the user request. For \ufo, it may mean that the planner never
created a subtask for a required result, or that a device agent used
\texttt{FINISH} even after explicitly reporting that the result had not
been created, after which the upper-level system treated the node as
complete. \ufo's D1 proportion reaches 20.4\%, substantially higher
than those of the other baselines. This shows that explicit task
decomposition does not by itself guarantee that the decomposition
covers every task requirement.

\paragraph{Confusion about device roles or target location (D2).}
In these trajectories, the agent understands what result must be
produced but writes it to the wrong device of the same type, the wrong
virtual machine, the wrong directory, or the wrong application
location. For example, an output may be required on the second Android
phone, but the model continues using the first phone, which served as
the information source. Alternatively, the source file may be on
\texttt{linux\_0} and the result may be required on \texttt{linux\_1},
but the model creates the result on the recently accessed source
device.

These errors show that although device IDs appear in the instruction
and observations, the agent does not consistently convert them into
persistent semantic roles such as information-source device,
target-output device, or final-verification device. As the trajectory
becomes longer, the most recently used device can replace the
originally specified target in the model's working state. In
hierarchical systems, the error may already occur when the planner
assigns the target device, after which the device agent simply executes
on the incorrectly assigned target. Claude Opus 4.8 has the highest D2
proportion, at 11.0\%.

\paragraph{Incorrect final output or device state (D3).}
In this category, the target file, application record, scheduled task,
or device state actually exists, but its fields, values, structure,
parameters, or action semantics do not satisfy the task requirements.
The model may read the correct source value and select the correct
device, yet alter the intended representation during information
transfer or result generation. Examples include using an incorrect JSON
nesting structure, omitting spreadsheet fields, creating an incomplete
scheduled task, or applying a SmartHome operation that is semantically
similar to but not identical to the required target state.

D3 accounts for between 40.9\% and 58.0\% of incorrect-completion
trajectories across the five baselines, indicating that precise
realization of the final state is a common cross-model bottleneck.
Agents frequently verify only that a file exists, a command returned
successfully, or a scheduled task was created, without checking whether
the internal content, parameters, and state are correct. For
hierarchical agents, exact source values may also be paraphrased as
they pass through the planner summary, subtask payload, and
device-agent output, causing the loss of precise values, field names,
or target formats.

These failures are therefore not cases in which the agent simply failed
to use a tool. Rather, the task requirements were not reliably
translated into an exact, persistent environment state that could pass
verification.

\paragraph{Declaring completion without confirming result persistence (D4).}
In these trajectories, the agent attempts a create, save, send, open,
or submit action but invokes the completion action without using a new
observation to confirm that the result was actually persisted. The
final verifier then finds that the target file, message, browser state,
or application record does not exist.

The central problem is that the agent conflates three different states:
an action has been issued, the interface appears plausible, and the
task result actually exists. Clicking a save button does not
necessarily mean that a file has been written. Clicking send does not
guarantee that the message appears in the target thread. Reading a
webpage source file does not mean that the browser has opened the
required page. Current agents lack a stable habit of reopening,
reading, or querying a result after an important write operation.
GPT-5.5 has the highest D4 proportion, at 19.7\%, indicating that it is
comparatively likely to treat a seemingly successful local action as
sufficient evidence that the result has been persisted.

\paragraph{Incomplete satisfaction of final conditions across multiple
devices (D5).}
In these tasks, the agent has operated in all relevant device
environments, but required conditions on at least two devices remain
unsatisfied when the task ends. For example, the SmartHome state may
have been correctly modified while the Android reply and Linux record
remain incorrect. Alternatively, Android, Linux, and SmartHome may each
contain a partial result, but the target phone for the note, the
scheduled-task settings, and the JSON fields do not all pass
verification at the same time.

This category most directly reflects the global-completion problem in
cross-device tasks. Models often inspect only the most recently
completed local operation or interpret the local \texttt{FINISH}
signals of several device agents as evidence that the overall task is
complete. They do not maintain a final checklist that covers all
postconditions across all devices. In other words, the fact that
several local subtasks have individually terminated does not imply that
they jointly form the correct cross-device final state.

GPT-5.5 and Gemini-3.1-Pro-Preview have D5 proportions of 30.3\% and
29.4\%, respectively, which are higher than those of the other
baselines. \ufo has a comparatively lower D5 proportion, but its D1 and
D2 proportions are higher. This does not justify the conclusion that a
hierarchical architecture provides stronger global-verification
capability. A more accurate observation is that \ufo and direct GPT-5.5
exhibit different failure-type distributions while retaining similar
overall success rates. Because the two systems differ in prompts,
history representation, action interfaces, and orchestration, this
comparison should be interpreted as descriptive rather than as a
controlled ablation of planner architecture.

\subsubsection{Local Progress Does Not Imply End-to-End Completion}
\label{subsubsec:local_progress}

The two termination modes and their associated failure types ultimately
point to a common issue: current agents can often perform useful local
operations but struggle to continuously maintain the complete state of
a cross-device task. Approximately 28.7\% of failed runs satisfy at
least one score-enabled condition. This means that a substantial
proportion of failures occur after the agent has made partial progress,
rather than after producing no valid result.

An agent may correctly read source information, generate a file, or
complete a SmartHome operation, but still fail because another output
is missing, the result was written to the wrong device, the structure
is incomplete, the state was not persisted, or the final checks were
insufficient. A reliable cross-device agent must continuously maintain
which pieces of critical information have already been obtained and
from which devices; the role of each device in the task; which subgoals
have been completed and which remain incomplete; whether existing
results have actually been persisted; and whether the final conditions
across all devices have been jointly satisfied.

At the same time, we do not interpret every failure simply as a
coordination failure. Failure to acquire required source information,
failure of a GUI output operation on a single device, file-format
errors, and result-persistence errors may initially be device-level
execution problems. What \textsc{DevicesWorld} reveals more clearly is
that these local problems propagate along cross-device dependencies,
while device-role binding, information transfer, and multi-endpoint
final-result verification introduce additional failure surfaces. The
central limitation of current agents is their inability to reliably
integrate local progress into verified, end-to-end cross-device task
completion.
\section{Discussion}
The experimental results and trajectory analysis on \textsc{DevicesWorld} show that current LLM agents remain far from reliable cross-device execution. Some failures occur because models remain stuck in local operations for long periods and cannot adjust their strategy based on environmental feedback. Other failures occur after the model has already made partial progress—for example, after reading source information, generating a file, or completing an operation on one device—but ends the task while other results are still missing or incorrect. Together, these phenomena show that current agents cannot reliably maintain the overall execution state of a cross-device task. Not every failure should be attributed to an abstract lack of coordination capability: device-level problems such as interface localization, application navigation, and file generation can also cause failures. In cross-device tasks, however, a local problem often propagates through dependencies between devices, blocking subsequent operations or corrupting the final result.

Future reliable cross-device agents should first maintain an explicit cross-device task state rather than relying only on an ever-growing interaction history. This state should record the role of each device in the task, the critical information already acquired, the source and target location of each piece of information, dependencies between devices, and the subgoals that remain incomplete. For example, an agent should not merely remember a time or a contact. It should also know which device provided that information, which subsequent operation will use it, and which device should ultimately receive the result. Such a representation would help reduce information loss after device switching and lower the probability of writing output to the wrong device or location.

Second, agents need dynamic replanning capabilities based on execution feedback and task dependencies, consistent with prior work on feedback-conditioned reasoning, reflection, and hierarchical replanning
~\cite{yao2023react,shinn2023reflexion,agashe2025agents2}. The trajectories show that current models frequently continue using the same or similar interaction strategy even after observing empty search results, an interface that remains unchanged for many steps, or an explicit action failure. A more reliable system should determine from execution feedback whether the current plan remains valid. When several consecutive steps fail to produce new critical information, complete a new subgoal, or resolve a repeated error, the agent should replan in time by changing the search method, switching the operation path, or selecting a different target device. For subgoals with prerequisite dependencies, the agent should not bypass an unsatisfied prerequisite. It should attempt alternative feasible paths for obtaining the required information and, when necessary, determine whether the condition cannot be satisfied. For subgoals that are independent of the current bottleneck, the agent may reorder execution and prioritize components that can still make progress. 

Finally, before confirming task completion, the agent must perform a global review of all relevant devices and outputs. Current models often inspect only the most recently completed local result and fail to verify whether the conditions on other devices have also been satisfied. A more reliable system should maintain a complete checklist derived from the user goal and confirm, item by item, that each output is located on the correct device, every required result exists, the content and structure are correct, device-state changes have actually taken effect, and results on different devices are mutually consistent. If any result remains missing or incorrect, the agent should return to the relevant device and repair it rather than terminating the task. Task verification and termination have also been identified as
distinct failure surfaces in agent systems, motivating explicit pre-termination checks over all task postconditions
~\cite{cemri2025masfail,lu2025agentrewardbench,
wei2026opencomputer}.

These three capabilities form an interconnected execution loop. Structured cross-device task state provides the basis for subsequent decisions and dynamic replanning. Replanning based on execution feedback and task dependencies allows the agent to adjust its information-acquisition strategy, operation path, or subgoal order when the current path fails. Global result verification ensures that local outputs on different devices ultimately satisfy the task requirements together. These capabilities may also reduce repeated searches, ineffective operations, and incorrect completion decisions. Future improvements to cross-device agents should not rely solely on longer context windows or larger interaction budgets. Instead, cross-device state maintenance, dynamic replanning, and task-completion verification should be integrated into a unified end-to-end execution process.

\section{Related Work}

\subsection{Executable Mobile, Web, and Desktop Agent Benchmarks}

A broad range of benchmarks has been developed to evaluate agents in
mobile, web, and desktop environments. On mobile devices, Android in
the Wild provides large-scale human demonstrations of Android device
control, B-MoCA evaluates generalization across different device
configurations, AndroidWorld provides dynamically instantiated
executable tasks with programmatic state-based evaluation, and GUI
Odyssey studies cross-application navigation within mobile devices
~\cite{rawles2023androidwildlargescaledataset,lee2025benchmarkingmobiledevicecontrol,rawles2025androidworld,
lu2025guiodysseycomprehensivedatasetcrossapp}.

Web-agent benchmarks cover realistic websites, visually grounded
interaction, enterprise workflows, and compositional knowledge-work
tasks. Mind2Web provides demonstrations collected from real-world
websites, WebArena evaluates long-horizon tasks in self-hosted
functional websites, VisualWebArena emphasizes visually grounded web
interaction, and WorkArena and WorkArena++ focus on enterprise
knowledge-work workflows
~\cite{deng2023mind2web,zhou2024webarenarealisticwebenvironment,
koh2024visualwebarena,drouin2024workarena,
boisvert2024workarenapp}. BrowserGym further provides standardized
observation, action, and evaluation interfaces for integrating
multiple web-agent benchmarks~\cite{chezelles2025the}.

Desktop benchmarks extend executable evaluation to full operating
systems. OSWorld evaluates open-ended tasks involving desktop
applications, file systems, browsers, and command-line tools;
Windows Agent Arena provides scalable evaluation in a Windows
environment; and WindowsWorld focuses on professional
cross-application workflows
~\cite{xie2024osworld,bonatti2024windowsarena,
li2026windowsworld}.

Most of these benchmarks still center on a single dominant execution
environment. Even when a task spans multiple applications, the agent
normally operates within one mobile device, browser environment, or
operating-system instance. \textsc{DevicesWorld} builds on their executable
task setup and state-based evaluation paradigms, but focuses on tasks
in which information sources, target operations, and final
postconditions may be distributed across multiple concrete devices.

\subsection{Cross-Device Interaction and Cross-Environment Agents}

Cross-device interaction has long been studied in human--computer
interaction. Prior work has identified challenges arising from
heterogeneous device capabilities, dynamically changing device
availability, and the allocation of information and interface
components across devices
~\cite{grubert2016multidevice,park2018adam}. These challenges remain
relevant when the actor coordinating the devices is an autonomous
agent rather than a human user.

More recently, agent research has begun to broaden evaluation beyond
a single environment. CRAB supports cross-environment tasks and
introduces graph-based evaluation for measuring fine-grained task
completion~\cite{xu2024crab}. MMBench-GUI evaluates GUI-agent
capabilities across Windows, macOS, Linux, iOS, Android, and Web,
including a Task Collaboration level and an efficiency-oriented
metric~\cite{wang2025mmbenchgui}. \ufo studies distributed
orchestration across heterogeneous endpoints by representing a user
request as a dynamically updated task graph with explicit data and
control dependencies~\cite{zhang2025ufo3}.

These studies establish important foundations for cross-environment
evaluation, multi-platform generalization, and distributed agent
orchestration. \textsc{DevicesWorld} addresses a complementary benchmark
setting: it provides 6,140 executable end-user tasks that jointly
involve mobile, desktop, and smart-home environments, supports
multiple concrete devices of the same or different environment
classes, and evaluates task completion through joint postconditions
over device states, application records, files, and other generated
artifacts.

\subsection{Smart-Home and IoT Agent Benchmarks}

Smart-home benchmarks evaluate user-intent understanding,
device-state reasoning, constraint handling, temporal planning, and
automation execution in household environments. HomeBench considers
both valid and invalid instructions involving single or multiple
smart-home devices~\cite{li2025homebench}. SimuHome introduces an
executable and time-accelerated environment in which device
operations affect environmental variables and agents must handle
state queries, temporal dependencies, and scheduling
~\cite{seo2025simuhome}. SMH-Bench evaluates state querying,
automation, ambiguity handling, and personalized reasoning in
executable homes of different complexity
~\cite{li2026smhbench}. HomeFlow further couples smart-home task and
trajectory generation with executable state-based success conditions
~\cite{gu2026homeflowt}.

These benchmarks demonstrate that coordination among multiple
devices can already be challenging within a single household
environment. \textsc{DevicesWorld} differs not merely by including additional
device categories, but by placing SmartHome within a broader
personal-device ecology. In \textsc{DevicesWorld}, the information or policy
governing a smart-home operation may originate from a mobile
application or a desktop file, while the required final results may
need to appear jointly in mobile applications, desktop artifacts, and
smart-home states. The evaluation target is therefore end-to-end task
execution across heterogeneous devices rather than device control
within the home alone.

\section{Conclusion}
The operational capabilities of LLM agents in individual environments such as mobile applications, desktop systems, and smart homes continue to improve, but real-world user tasks often require several devices to participate jointly. There remains a lack of large-scale executable evaluation infrastructure for systematically assessing whether an agent can continuously acquire information, perform operations, and complete a single user goal across heterogeneous devices. To address this gap, we introduce \textsc{DevicesWorld}, a large-scale executable benchmark for cross-device collaborative operation. \textsc{DevicesWorld} contains 6,140 tasks and a unified cross-device interaction and evaluation environment covering three classes of device environments: mobile, desktop, and IoT. It supports tasks involving multiple devices within the same environment class as well as tasks jointly involving heterogeneous device environments. Each task defines a natural-language user goal, participating devices, initial states, executable interactions, result verification, and environment cleanup. A multi-stage task-construction and quality-control pipeline ensures both executability and consistency between task requirements and verification conditions. We evaluate five representative LLM-agent baselines on \textsc{DevicesWorld}, with the highest task success rate reaching only 12.5\%, demonstrating that current agents remain far from reliable cross-device execution. Trajectory analysis further shows that some failed runs remain stuck in local search or operation until the interaction budget is exhausted, while others declare completion before the full set of task conditions has been satisfied. More generally, even when current agents make local progress, they struggle to continuously maintain device roles, cross-device dependencies, task progress, and complete completion conditions, and to integrate local outcomes into verified end-to-end task completion. These results indicate that a reliable cross-device agent requires more than stronger local-operation capability. It must explicitly maintain cross-device task state, dynamically replan according to execution feedback and task dependencies when the current path fails, and verify all relevant devices and outputs before confirming task completion. \textsc{DevicesWorld} provides an executable and diagnostically useful foundation for systematically evaluating these capabilities, analyzing failures in cross-device execution, and advancing research on reliable cross-device agents.

\bibliography{iclr2026_conference}
\bibliographystyle{iclr2026_conference}

\end{document}